\documentclass[11pt]{article}
\usepackage{rldmsubmit,palatino}
\usepackage{graphicx}
\usepackage{natbib}
\usepackage{wrapfig}
\usepackage{amsmath}
\usepackage{amssymb}
\usepackage{cleveref}
\title{Scaling CrossQ with Weight Normalization}

\author{
Daniel Palenicek \\
Technical University of Darmstadt \\
hessian.AI \\
\texttt{palenicek@robot-learning.de} \\
\And
Florian Vogt \\
University of Freiburg \\
\texttt{vogtf@informatik.uni-freiburg.de} \\
\And
Jan Peters \\
Technical University of  Darmstadt \\
hessian.AI \\
German Research Center for AI (DFKI) \\
Robotics Institute Germany (RIG) \\
\texttt{jan@robot-learning.de}
}

\newcommand{\RL}{\textsc{rl}}
\newcommand{\ELR}{\textsc{elr}}
\newcommand{\WN}{\textsc{wn}}
\newcommand{\DMC}{\textsc{dmc}}
\newcommand{\UTD}{\textsc{utd}}
\newcommand{\BN}{\textsc{bn}}
\newcommand{\LN}{\textsc{ln}}
\newcommand{\MDP}{\textsc{mdp}}
\newcommand{\IQM}{\textsc{iqm}}

\newcommand{\SAC}{\textsc{sac}}
\newcommand{\BRO}{\textsc{bro}}

\newcommand{\mujoco}{\texttt{MuJoCo}}

\usepackage{bm}
\newcommand{\vs}{\bm{s}}
\newcommand{\va}{\bm{a}}

\begin{document}

\maketitle

\begin{abstract}

Reinforcement learning has achieved significant milestones, but sample efficiency remains a bottleneck for real-world applications.
Recently, CrossQ has demonstrated state-of-the-art sample efficiency with a low update-to-data (\UTD{}) ratio of 1.
In this work, we explore CrossQ's scaling behavior with higher \UTD{} ratios.
We identify challenges in the training dynamics which are emphasized by higher \UTD{}s, particularly Q-bias explosion and the growing magnitude of critic network weights.
To address this, we integrate weight normalization into the CrossQ framework, a solution that stabilizes training, prevents potential loss of plasticity and keeps the effective learning rate constant.
Our proposed approach reliably scales with increasing \UTD{} ratios, achieving competitive or superior performance across a range of challenging tasks on the DeepMind control benchmark, notably the complex \texttt{dog} and \texttt{humanoid} environments.
This work eliminates the need for drastic interventions, such as network resets, and offers a robust pathway for improving sample efficiency and scalability in model-free reinforcement learning.
\end{abstract}


\acknowledgements{This research funded by the Research Clusters “Third Wave of AI” and “The Adaptive Mind”, funded by the Excellence Program of the Hessian Ministry of Higher Education, Science, Research and the Arts, hessian.AI.}

\startmain

\section{Introduction}

Reinforcement Learning (\RL{}) has shown great successes in recent years.
One fundamental question in \RL{} that remains is how to increase the sample efficiency of the algorithms.
This is one crucial part for making \RL{} methods applicable to real-world scenarios like robotics.
To that end, model-based \RL{} attempts to increase sample efficiency by learning dynamic models that reduce the need for collecting real data, a process often expensive and time-consuming~\citep{sutton1990dynaQ,janner2019mbpo,cowen2022samba}.
Contrary to this approach, there have been attempts to increase the sample efficiency of purely model-free algorithms.
Either by increasing the number of gradient updates on the available data, also referred to as update-to-data (\UTD{}) ratio~\citep{chen2021redq,hiraoka2021droq,nikishin2022primacy,doro2022replaybarrier}, changes to the network architectures~\citep{bhatt2024crossq}, or both~\citep{hussing2024dissecting,nauman2024bigger}.

In this work, we build upon CrossQ~\citep{bhatt2024crossq}, a model-free \RL{} algorithm that recently showed state-of-the-art sample efficiency on the \mujoco{}~\citep{todorov2012mujoco} continuous control benchmarking tasks.
Notably, the authors achieved this by carefully utilizing Batch Normalization (\BN{}) within the actor-critic architecture.
A technique that has previously been thought to not work in \RL{} as famously reported by~\citet{hiraoka2021droq} and others.
The insight that \citet{bhatt2024crossq} offered, is that one needs to carefully consider the different state-action distributions within the Bellman equation and handle them correctly to succeed.
This novelty allowed maintaining a low \UTD{} ratio of 1 gradient step per environment interaction, while outperforming then state-of-the-art algorithms that scaled the \UTD{} ratio to 20.

This naturally raises the question: \textit{How can we extend the sample efficiency benefits of CrossQ and \BN{} to the high \UTD{} training regime?} Which we address in this manuscript.

\paragraph{Contributions.}
In this work, we show, that the vanilla CrossQ algorithm can be brittle to tune on DeepMind Control~(\DMC{}) environments.
We further show that it does not scale reliably with increasing compute.
We propose the addition of weight normalization~(\WN{}).
We demonstrate, that \WN{} is an easy but very effective measure to stabilize CrossQ.
Further, we find that these stabilizing effects allow to scale the~\UTD{} of CrossQ and consequently its sample efficiency.

\section{Preliminaries}

\paragraph{Reinforcement learning.}
Consider a Markov Decision Process (\MDP{})~\citep{puterman2014mdp}, which is defined by the tuple  $\mathcal{M} = \langle\mathcal{S}, \mathcal{A}, \mathcal{P}, r, \gamma\rangle$, with the state space $\mathcal{S} \subseteq \mathbb{R}^n$, action space $\mathcal{A} \subseteq \mathbb{R}^m$, transition probability $\mathcal{P}: \mathcal{S}\times\mathcal{A}\rightarrow{}\Delta(\mathcal{S})$, the reward function $r:\mathcal{S}\times\mathcal{A}\rightarrow{}\mathbb{R}$ and discount factor $\gamma$.
We define the \RL{} problem according to \citet{Sutton1998}.
A policy $\pi:\mathcal{S}\rightarrow{}\Delta(\mathcal{A})$ is a behavior plan, which maps a state to a distribution over actions.
The goal of \RL{} is to find an optimal policy $\pi^*$ that maximizes the discounted cumulative reward
$\mathcal{R}_t = \sum_{k=t}^{\infty}\gamma^{k-t} r(\vs_k, \va_k)$, also referred to as the \textit{return}.
The Q-function
$Q^\pi(\vs,\va) = \mathbb{E}_{\pi,\mathcal{P}} [\mathcal{R}_t | \vs_t=\vs, \va_t=\va ]$ defines the expected return when taking action $\va$ in state $\vs$.

For a detailed introduction to CrossQ, we refer the reader to \citet{bhatt2024crossq}, due to space constraints.
In a nutshell, CrossQ introduces \BN{} into the critic (and optionally actor) networks, widens the layers and removes target networks.

\paragraph{Normalization techniques in \RL{}.}
Normalization techniques are widely recognized for improving the training of neural networks, as they generally accelerate training and improve generalization~\citep{huang2020normalizationtechniquestrainingdnns}.
There are many ways of introducing different types of normalizations into the \RL{} framework.
Most commonly, authors have used Layer Normalization~(\LN{}) within the network architectures to stabilize training~\citep{hiraoka2021droq,lyle2024normalization}.
Recently, CrossQ has been the first algorithm to successfully use \BN{} layers in \RL{}~\citep{bhatt2024crossq}.
The addition of \BN{} lead to substantial gains in sample efficiency.
In contrast to \LN{}, however, one needs to carefully consider the different state-action distributions within the critic loss when integrating \BN{}.
In a different line of work, \citet{hussing2024dissecting} proposed to project the output features of the penultimate layer onto the unitball to fight Q-function overestimation.

\citet{van2017l2} noted that normalization techniques such as \BN{} and \LN{} yield functions $f$ that are scale invariant with respect to the weights $\bm{w}$, i.e., $f(\bm{X};c\bm{w},\gamma,\beta) =  f(\bm{X};\bm{w},\gamma,\beta)$.
Further, the gradients scale inversely proportional to the scaling factor $\nabla f(\bm{X};c\bm{w},\gamma,\beta) =  \nabla f(\bm{X};\bm{w},\gamma,\beta) / c$.
In deep learning, one commonly observes the magnitudes of the network parameters to grow larger over the course of training.
Together with inversely proportional scaling of the gradients, this induces an implicit decreasing learning rate schedule.
\citet{lyle2024normalization} recently investigated the effectiveness of combining \LN{} and \WN{} to keep the effective learning rate~(\ELR{})~\citep{van2017l2} constant in order to avoid loss of plasticity in discrete state-action \RL{}.

\paragraph{Increasing update-to-data ratios.}
Although scaling up the \UTD{} ratio is an intuitive approach to increase the sample efficiency, in practice, it comes with several challenges.
\citet{nikishin2022primacy} demonstrated that overfitting on early training data can inhibit the agent from learn anything later in the training. The authors dub this phenomenon the primacy bias.
They suggest to periodically reset the network parameterized while retraining the replay buffer.
Many works that followed have adapted this intervention~\citep{doro2022replaybarrier,nauman2024bigger}.
While often effective, regularly resetting is a very drastic intervention and by design induces regular drops in performance.
Since the agent has to repeatedly start learning from scratch, it is also not very compute efficient.
Finally, the exact reasons why parameter resets work so well in practice are not well understood~\citep{li2023efficientdeeprl}.
Instead of resetting there have also been other types of regularization that allowed practitioners to train stably with high \UTD{} ratios.
\citet{janner2019mbpo} generates additional modeled data, by virtually increasing the \UTD{}. In \textsc{redq}, \citet{chen2021redq} leverage ensembles of Q-functions, while \citet{hiraoka2021droq} use dropout and \LN{} to effectively scale to higher \UTD{} ratios.

\section{Failure to Scale Stably}
\begin{figure}[t]
    \centering
    \includegraphics[width=\linewidth]{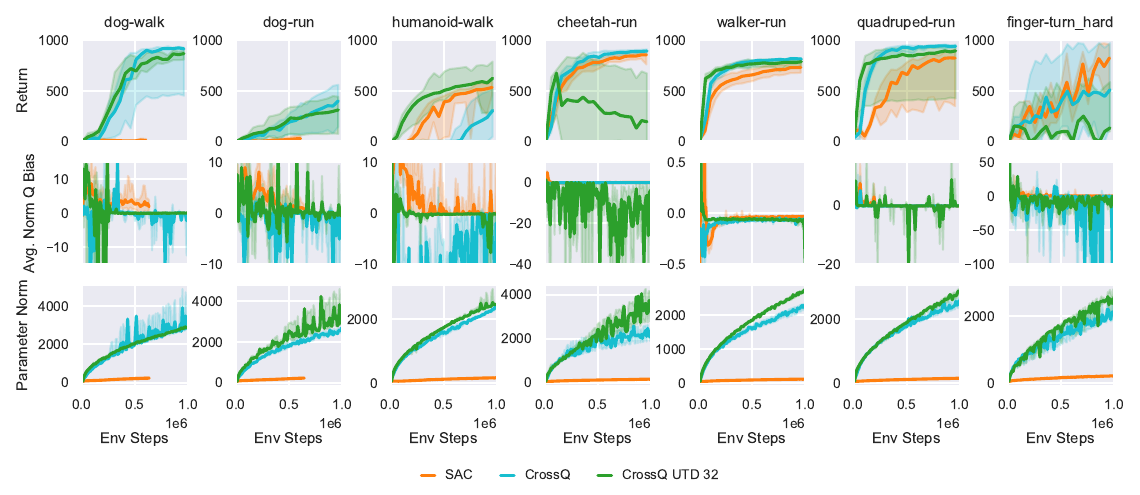}
    \vspace{-2em}
    \caption{\textbf{Q-bias and weight norms.}
    In general CrossQ shows much larger weight norms than \SAC{}. High \UTD{} can emphasize this problem even more.
    Large and high variance Q-bias can often be linked to poor performance.
    }
    \label{fig:crossQ-failure}
\end{figure}

\citet{bhatt2024crossq} demonstrated CrossQ's state-of-the-art sample efficiency on the original \mujoco{} task suite~\citep{todorov2012mujoco}, while at the same time, being very computationally efficient.
However, on the more extensive \DMC{} task suite, we find that CrossQ can require tuning.
We further find that it works on some, but not all environments stably and reliably.

\Cref{fig:crossQ-failure} shows CrossQ training performance on a subset of \DMC{} tasks.
We picked the tasks to show a diverse set of behaviors, from successful learning to instabilities during learning.
The results on the remaining tasks are similar but were left out due to space constraints.
The second and third rows show the Q-function bias and the norm of the network weight matrices of the critic network, respectively.
On the one hand, vanilla CrossQ can show instabilities during training or trains slowly, if at all.
On the other hand, high \UTD{} ratios can emphasize this effect in some environments.
We can link bad performance of CrossQ to fluctuating and exploding Q-bias values.
Which in turn have a connection to the common issue of Q-function overestimation.

Further, we notice the growing magnitude of the critic weight matrices.
While all three baselines show growing network weights, the effect is vastly emphasized for CrossQ and even more with higher \UTD{} ratios.
Growing network weights are a phenomenon which has been linked to loss of plasticity, and can therefore, lead to premature convergence.
Additionally, the growing magnitudes poses a challenge for the optimization in itself, similar to the issue of growing activations, the latter of which has been analyzed by \citet{hussing2024dissecting}.

\section{Successfully Scaling CrossQ to High Update-To-Data Ratios with Weight Normalization}

As we demonstrate in the previous section and reported in many other works~\citep{chen2021redq,hiraoka2021droq}, the Q-bias can be linked to poor training performance.
Particularly, in high \UTD{} settings the Q-bias tends to increase and explode if no countermeasures are taken.
Inspired by the combined insights of \citet{hussing2024dissecting} and \citet{lyle2024normalization}
we propose to combine CrossQ with \WN{} as a means of counteracting Q-bias explosion and instabilities of high \UTD{} ratios.
Our reasoning goes is as follows:
Due to the use of \BN{} in CrossQ, the critic network is scale invariant, as noted by~\citet{van2017l2}.
Therefore, we can employ \WN{} to counteract the growth of the networks weight norms.
\citet{lyle2024normalization} showed, that for \LN{} + \WN{}, this helps against loss of plasticity.
The scaling of the output is now left to the very last layer of the network only, similar to the unit ball normalization of the penultimate layers outputs proposed by~\citet{hussing2024dissecting}.
And the \ELR{} is kept constant.

\begin{wrapfigure}[23]{r}{2.1in}
    \vspace{-2.5em}
    \begin{center}
    \includegraphics[width=2in]{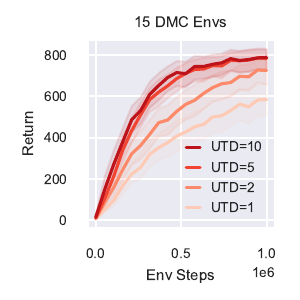}
  \end{center}
      \caption{\textbf{CrossQ \WN{} \UTD{} scaling behavior.} We plot the \IQM{} return and $95\%$ confidence intervals for different \UTD{} ratios $\in\{1,2,5,10\}$.
      The results are aggregated over 15 \DMC{} environments and $10$ random seeds each according to~\citet{agarwal2021iqm}.
      The sample efficiency scales reliably with increasing \UTD{} ratios.
    }
    \label{fig:utd-scaling}
\end{wrapfigure}

\paragraph{Weight normalization allows CrossQ to scale effectively.}
\begin{figure}[t]
    \centering
    \includegraphics[width=\linewidth]{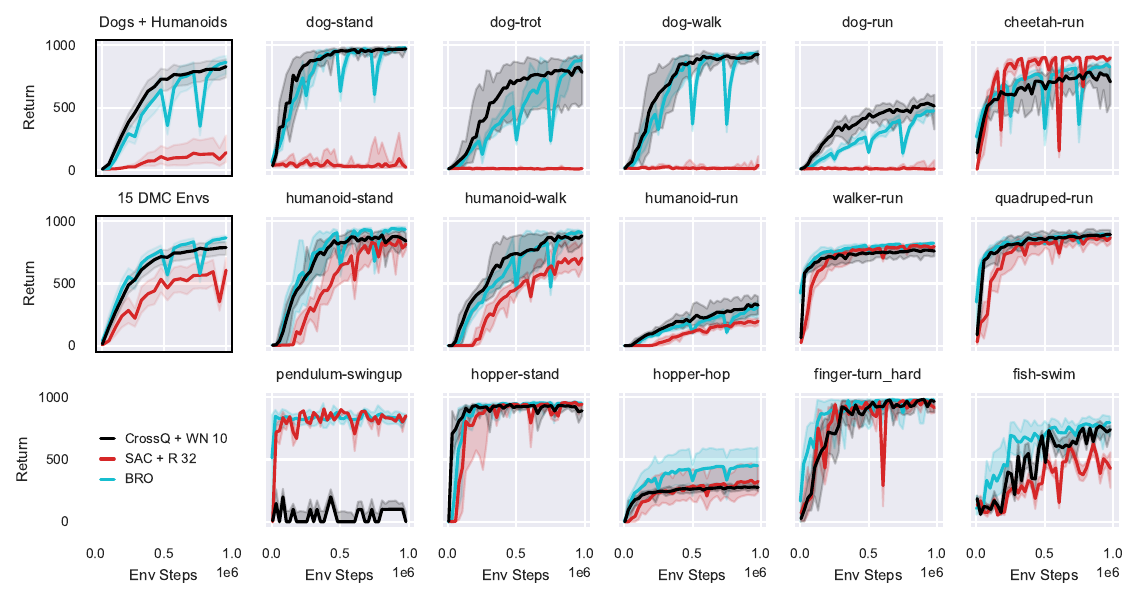}
    \caption{\textbf{CrossQ \WN{} \UTD{}=10 against baselines.}
    We compare our proposed CrossQ \WN{} \UTD{}=10 against two baselines, \BRO{}~\citep{nauman2024bigger} and \SAC{} with \UTD{}=32 and resets.
    Results are reported on all 15 \DMC{} tasks and the \IQM and $90\%$ inter percentile ranges over 10 random seeds.
    The leftmost plots (black border) provide aggregated results over the dog and humanoid environments and over 15 environments, respectively.
    Our proposed approach proves competitive to \BRO{} and outperforms the \SAC{} baseline.
    We want to note, that our approach achieces this performance without requiring resetting and is also much simpler than \BRO{}.
    }
    \label{fig:crossQ_wn_comparison}
\end{figure}

Next, we provide empirical evidence for our hypothesis.
We show that the addition of \WN{} helps to stabilize CrossQ and allows to stably increase the \UTD{} ratio.
On 15 \DMC{} tasks, we compare our proposed CrossQ + \WN{} to the recently proposed \BRO{} algorithm~\citep{nauman2024bigger} and \SAC{} with a \UTD{}=32 and periodic resetting.
For computational reasons, for the \BRO{} baseline we plot the official training data the authors published.

\Cref{fig:crossQ_wn_comparison} shows the results of our experiments.
THe leftmost column shows aggregated performances over the \texttt{dog} and \texttt{humanoid} environments, and over all 15 \DMC{} environments respectively.
The aggregation was performed according to~\citet{agarwal2021iqm}.
These results show, that CrossQ + \WN{} $\text{\UTD}=10$ is competitive to the \BRO{} baseline, especially on the more complex \texttt{dog} and \texttt{humanoid} tasks.
For individual environemnts, we plot the IQM return and $90\%$ inter percentile ranges across $10$ seeds each.
Again, we observe that our proposed CrossQ + \WN{} $\text{\UTD}=10$ trains very stably.
The combination with \WN{} manages to eliminate the stability problems of the base algorithm.
In most environments, it can compete with the much more complex \BRO{} baseline, with the exception of \texttt{hopper-hop} and \texttt{pendulum-swingup}.

For \texttt{hopper-hop}, it seems to be environment specific, as \citet{bhatt2024crossq} also did not achieve good results on the \mujoco{} \texttt{Hopper-v4}.
For \texttt{pendulum-swingup}, we can attribute the low performance to the task's sparse reward. We leave investigation of this to future work.
More interestingly, CrossQ + \WN{} \UTD{}=10 matches or even outperforms \BRO{} on the majority of the most difficult \texttt{humanoid} and \texttt{dog} tasks.
Further, it also does not suffer from the periodic drops in performance caused by frequent network resetting.
This leads us to conclude, that CrossQ benefits from the addition of \WN{}, which results in stable training and scales well with high \UTD{} ratios.
The resulting algorithm can match or outperform state-of-the-art baselines on the continuous control \DMC{} suite.

\paragraph{Stable scaling of CrossQ + \WN{} across different \UTD{} ratios.}
This ablations shows the stable scaling behavior of our proposed CrossQ + \WN{} $\text{\UTD}=10$ with increasing \UTD{} ratios.
\Cref{fig:utd-scaling} provides aggregated training curves $\text{\UTD{}}\in\{1,2,5,10\}$ aggregated over all 15 \DMC{} tasks.
The learning curves are nicely ordered in accordance with their \UTD{} ratio.
The same is true for individual learning curves in the environments, however, due to space constraints, we do not include this figure here.

\section{Conclusion}
In this work, we have addressed the instability and scalability limitations of CrossQ in \RL{} by integrating \WN{}.
Our empirical results demonstrate that \WN{} effectively stabilizes training and allows CrossQ to scale reliably with higher \UTD{} ratios.
The proposed CrossQ + \WN{} approach achieves competitive or superior performance compared to state-of-the-art baselines across a diverse set of \DMC{} tasks, including the complex humanoid and dog environments.
By eliminating the need for drastic interventions such as network resets, this extension preserves simplicity while enhancing robustness and scalability.

\bibliography{bibliography}

\end{document}